\documentclass[letterpaper, 10 pt, conference]{ieeeconf}  

\IEEEoverridecommandlockouts                              

\usepackage{svg} 
\usepackage{multirow}
\usepackage{tabularx}
\usepackage{caption}
\usepackage{graphicx}
\usepackage{amsfonts}
\usepackage{algorithm} 
\usepackage{algpseudocode} 
\usepackage{mathrsfs}
\usepackage{hyperref}

\usepackage{tabularx}
\usepackage{tabulary}
\newcolumntype{Y}{>{\centering\arraybackslash}X}

\usepackage{hyperref}
\usepackage{nameref}

\usepackage{amsmath}
\usepackage{amsthm}
\usepackage{amssymb}
\usepackage{mathrsfs}
\usepackage{aligned-overset}

\usepackage{tikz}

\usepackage{cite}

\usepackage[font=footnotesize]{caption}
\usepackage[labelformat=simple]{subcaption}

\usepackage{booktabs}
\usepackage{multirow}

\makeatletter
               {\list{}{\leftmargin=0pt
                        \labelwidth\z@ \itemindent-\leftmargin
                        }}%
               {\endlist}
\makeatother

\usepackage[english]{babel}
\usepackage{amsmath,lipsum}
\usepackage{etoolbox}
\newcommand{\zerodisplayskips}{%
  \setlength{\abovedisplayskip}{4pt}%
  \setlength{\belowdisplayskip}{4pt}%
  \setlength{\abovedisplayshortskip}{2pt}%
  \setlength{\belowdisplayshortskip}{2pt}}
\appto{\normalsize}{\zerodisplayskips}
\appto{\small}{\zerodisplayskips}
\appto{\footnotesize}{\zerodisplayskips}

\makeatletter
\let\orgdescriptionlabel\descriptionlabel
\renewcommand*{\descriptionlabel}[1]{%
  \let\orglabel\label
  \let\label\@gobble
  \phantomsection
  \edef\@currentlabel{#1}%
  \let\label\orglabel
  \orgdescriptionlabel{#1}%
}
\makeatother

\makeatletter
\renewcommand*\env@matrix[1][0.5\arraystretch]{%
  \edef\arraystretch{#1}%
  \hskip -\arraycolsep
  \let\@ifnextchar\new@ifnextchar
  \array{*\c@MaxMatrixCols c}}
\makeatother

\newtheorem{theorem}{Theorem}

\newtheorem{example}[theorem]{Example}

\usepackage{graphics}

\usepackage[absolute,overlay]{textpos}

\usepackage{xargs}
\usepackage[colorinlistoftodos,prependcaption,textsize=scriptsize]{todonotes}
\newcommandx{\unsure}[2][1=]{\todo[linecolor=red,backgroundcolor=red!25,bordercolor=red,#1]{#2}}
\newcommandx{\change}[2][1=]{\todo[linecolor=blue,backgroundcolor=blue!25,bordercolor=blue,#1]{#2}}
\newcommandx{\info}[2][1=]{\todo[linecolor=OliveGreen,backgroundcolor=OliveGreen!25,bordercolor=OliveGreen,#1]{#2}}
\newcommandx{\improvement}[2][1=]{\todo[linecolor=Plum,backgroundcolor=Plum!25,bordercolor=Plum,#1]{#2}}
\newcommandx{\thiswillnotshow}[2][1=]{\todo[disable,#1]{#2}}

\newcommandx{\yukai}[2][1=]{\todo[linecolor=cyan,backgroundcolor=cyan!25,bordercolor=cyan,#1]{\textbf{Yu-Kai:} #2}}
\newcommandx{\yukaiWIP}[2][1=]{\todo[linecolor=red,backgroundcolor=red!25,bordercolor=red,#1]{\textbf{[WIP] Yu-Kai:} #2}}

\newcommand{\commentOut}[1]{}

\title{\LARGE \bf
 Multi-modal Motion Prediction using Temporal Ensembling \\ with Learning-based Aggregation
}

\author{Kai-Yin Hong, Chieh-Chih Wang and Wen-Chieh Lin
\thanks{Kai-Yin Hong is with the Institute of Electrical and Computer Engineering,
        National Yang Ming Chiao Tung University, Hsinchu, Taiwan. {\tt\footnotesize{kaiyin0208.ee11@nycu.edu.tw}}}   
\thanks{Chieh-Chih Wang
        is with the College of Electrical and Computer Engineering, National Yang Ming Chiao Tung University, and with the Mechanical and Mechatronics Systems Research Laboratories, Industrial Technology Research Institute, Hsinchu, Taiwan. 
        {\tt\footnotesize {bobwang@ieee.org}}}%
\thanks{Wen-Chieh Lin is with the Institute of Multimedia Engineering, National Yang Ming Chiao Tung University, Hsinchu, Taiwan 
        {\tt\footnotesize wclin@cs.nctu.edu.tw}}%
}

\begin{document}

\thispagestyle{empty}
\pagestyle{empty}

\maketitle

\begin{textblock*}{\textwidth}(.1\textwidth,1em)

\end{textblock*}

\begin{abstract}


Recent years have seen a shift towards learning-based methods for trajectory prediction, with challenges remaining in addressing uncertainty and capturing multi-modal distributions. This paper introduces \textit{Temporal Ensembling with Learning-based Aggregation}, a meta-algorithm designed to mitigate the issue of missing behaviors in trajectory prediction, 
which leads to inconsistent predictions across consecutive frames. Unlike conventional model ensembling, temporal ensembling leverages predictions from nearby frames to enhance spatial coverage and prediction diversity. By confirming predictions from multiple frames, temporal ensembling compensates for occasional errors in individual frame predictions. Furthermore, trajectory-level aggregation, often utilized in model ensembling, is insufficient for temporal ensembling due to a lack of consideration of traffic context and its tendency to assign candidate trajectories with incorrect driving behaviors to final predictions. We further emphasize the necessity of learning-based aggregation by utilizing mode queries within a DETR-like architecture for our temporal ensembling, leveraging the characteristics of predictions from nearby frames. Our method, validated on the Argoverse 2 dataset, shows notable improvements: a 4\% reduction in minADE, a 5\% decrease in minFDE, and a 1.16\% reduction in the miss rate compared to the strongest baseline, QCNet, highlighting its efficacy and potential in autonomous driving. 
\end{abstract}


\newcolumntype{C}[1]{>{\centering\arraybackslash}p{#1}}
\section{Introduction}
\label{chatper:intro}

Autonomous driving technology, since its inception in the 1980s with Pomerleau's groundbreaking work~\cite{pomerleau1988alvinn}, has profoundly impacted daily lives, particularly in ensuring safe and comfortable path planning and collision avoidance through accurate anticipation of surrounding traffic. Motion prediction, therefore, plays a crucial role in various autonomous driving applications, including risk estimation~\cite{lefevre2014survey}, decision making~\cite{zhan2018probabilistic}, and traffic simulation~\cite{barcelo2010fundamentals}.
In recent years, motion prediction has increasingly relied on learning-based methods~\cite{gao2020vectornet, liang2020learning, ngiam2021scene, varadarajan2022multipath++, zhou2022hivt, zhou2023query}. Despite significant progress in this field, substantial challenges persist, particularly in capturing multi-modality and dealing with the considerable uncertainty in output predictions, especially as the prediction horizon extends. For instance, when a vehicle approaches an intersection, its actions may vary depending on the driver's long-term goals, leading to multiple possible trajectories. Overcoming this challenge requires motion prediction models to learn and capture the underlying multi-modal distribution instead of predicting only the most common mode. However, this task is complex, as each training sample typically represents only one possibility.

\begin{figure}[t]
\centering
\includegraphics[height=3.4cm]{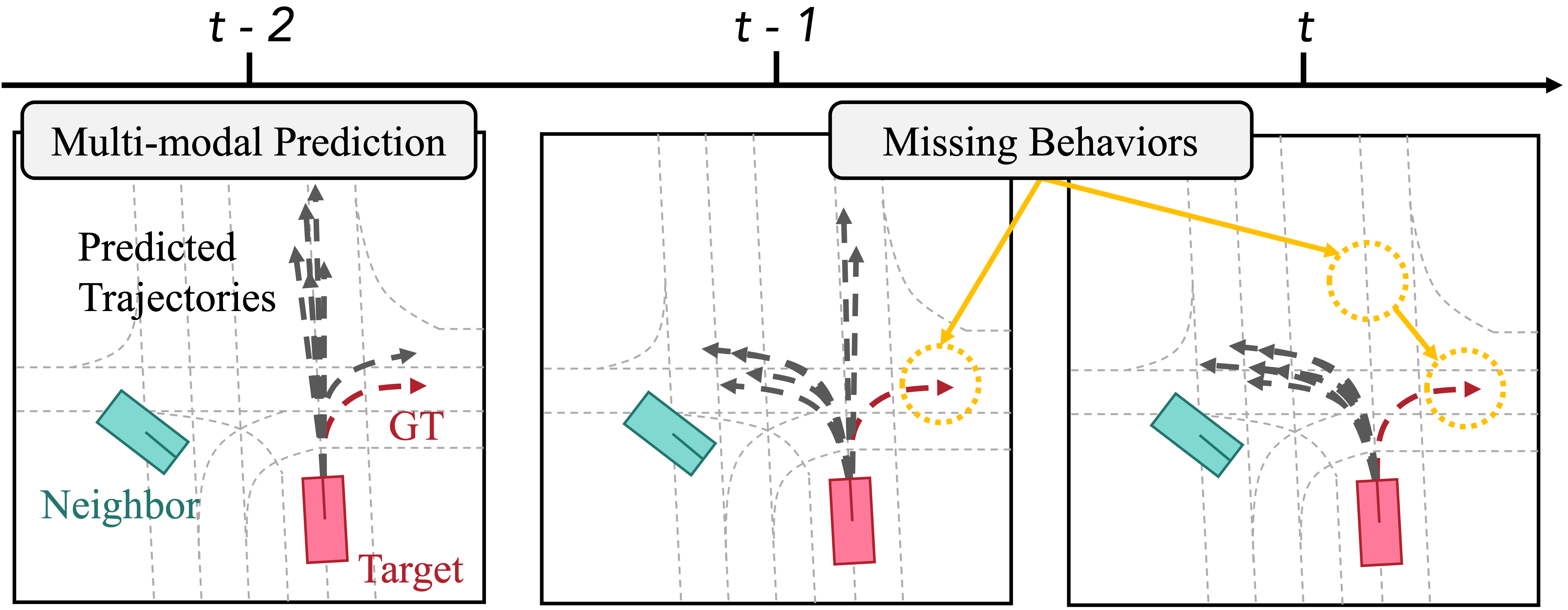}
\caption{Illustration of Missing Behaviors Issue - Missing behaviors refer to cases where predictions are occasionally wrong, resulting in inconsistent predictions across consecutive frames. The left panel depicts multi-modal motion prediction. The red car represents the target agent, with its red trajectory showing the ground truth. The gray trajectories illustrate various possible future paths. The middle and right panels demonstrate missing behaviors in the predicted trajectories across consecutive time steps.}
\vspace*{-0.6em}
\label{fig:issue}
\end{figure}

When applying the state-of-the-art approach, QCNet \cite{zhou2023query} to real-world scenarios, we observed instances where predictions may be inaccurate and fail to capture the exact behavior. For example, when examining the predicted trajectories at time step $t$ in Fig. \ref{fig:issue}, all predictions suggest a left turn, but the actual movement turns out to be a right turn. Furthermore, upon closer observation, we observed that the predictions at time step $t-2$ were accurate, despite the error at time step $t$. We aim to leverage this characteristic to address missing behaviors.

We first propose \textbf{Temporal Ensembling} to address missing behaviors. By integrating predictions from nearby frames, we aim to compensate for occasional errors in individual frame predictions, enhance spatial coverage, and improve accuracy. Similar to the model ensembling employed by \cite{zhou2023query, wang2023prophnet, varadarajan2022multipath++, shi2022motion} where multiple models are trained and their predictions combined to enhance performance, temporal ensembling accomplishes motion predictions with a single model but with predictions from multiple frames.

Temporal ensembling shares similarities with model ensembling by increasing the pool of prediction candidates before aggregating them into the final predictions. However, trajectory-level aggregation methods such as Top-K, Non-Maximum Suppression (NMS), and K-means - commonly used in model ensembling \cite{varadarajan2022multipath++,wang2023prophnet,zhou2023query} 
- are found to be inadequate for temporal ensembling. 
This is because trajectory-level aggregation does not consider traffic context and tends to incorrectly assign candidate trajectories,
which exhibit inappropriate driving behaviors, to the final predictions. This limitation motivates the need for a more dynamic and context-aware approach. We propose \textbf{Learning-based Aggregation} to enable temporal ensembling by utilizing mode queries within a DETR-like architecture\cite{carion2020end} to determine the final predictions.

The main contribution of this paper is introducing a meta-algorithm named Temporal Ensembling with Learning-based Aggregation to address missing behaviors. We validated our approach on the Argoverse 2 dataset\cite{wilson2021argoverse}, achieving notable improvements in the three crucial prediction metrics: a 4\% enhancement in minADE, a 5\% improvement in minFDE, and a 1.16\% reduction in miss rate compared to the strongest baseline, QCNet\cite{zhou2023query}. 
\section{Related Works}
\label{sec:related-works}

Predicting vehicle behavior is crucial for autonomous driving. Initially, research focused on physics-based models for short-term predictions, as highlighted in studies such as \cite{lefevre2014survey} and \cite{barth2008will}. Recently, the trend has shifted to learning-based methods, which are preferred for their accuracy and ability to account for road interactions~\cite{chen2022milestones}. 
We'll review the latest work in multi-modal motion prediction and trajectories ensembling techniques. For a complete review on motion prediction, please see \cite{chen2022milestones}.

\subsection{Multi-Modal Motion Prediction}
Due to the uncertainty in agent intent, motion prediction outputs naturally exhibit multiple modes, making it challenging to determine whether a vehicle will proceed straight or turn right as it approaches an intersection. Therefore, having a model capable of capturing multiple potential trajectories within a limited set is crucial. 
Several studies \cite{cui2019multimodal, kim2021lapred, huang2022multi} aim to predict multiple future trajectories and associated probabilities through direct regression. To enhance the coverage of potential outcomes, anchor-based approaches have gained popularity in trajectory prediction. These approaches first classify discrete intent and then regress continuous trajectories conditioned on the identified intent. Intent classification can be categorized into two types: goal-based (goal positions \cite{zhao2021tnt, gu2021densetnt} or target lanes \cite{kim2021lapred}) and driving maneuver-based \cite{deo2018convolutional}. The recent state-of-the-art method \cite{zhou2023query} utilizes the DETR-like \cite{carion2020end} architecture to address the multi-modal problem, combining anchor-free and anchor-based techniques to achieve notable performance. Despite advancements in multi-modal problems, a gap remains in achieving both accurate and comprehensively covered predictions. Our analysis of the existing model\cite{zhou2023query} identifies inherent limitations, termed missing behaviors, as illustrated in Fig. \ref{fig:issue}. This insight has led us to develop strategies to mitigate the multi-modal problem, focusing on tackling the issue of missing behaviors.

\subsection{Trajectories Ensembling Techniques}
\begin{figure}[t]
\includegraphics[height=2.4cm]{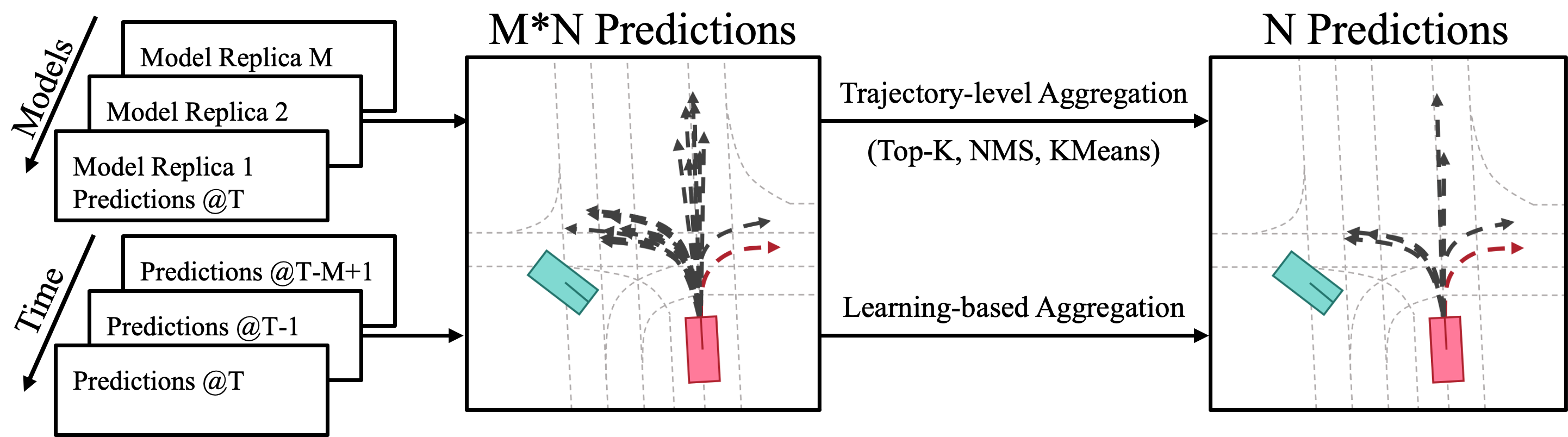}
\caption{Comparison of Ensembling Methods - The figure illustrates two ensembling approaches. \textbf{Model Ensembling (Top):} Multiple models independently predict N trajectories at Frame t. With M models, this results in M*N trajectories that are combined into the final N trajectories at the trajectory level. \textbf{Temporal Ensembling (Bottom):} A single model generates M*N predictions across M nearby frames. Our proposed learning-based aggregation then combines them into the final N trajectories.}
\label{fig:ensemble-comparison}
\vspace*{-0.6em}

\end{figure}

\label{related-works:ensemble}
Model ensembling \cite{ganaie2022ensemble} is a widely adopted technique to enhance motion prediction performance \cite{zhou2023query, wang2023prophnet, varadarajan2022multipath++, shi2022motion}. Typical strategies involve training multiple replicas of the model with varying initial parameters, learning rates, epochs, or data portions. 
In these model ensembling-based approaches, while a single model generates $N$ trajectories, employing $M$ models can yield $M*N$ trajectories.
However, post-processing is necessary to integrate outputs from multiple models for a fair comparison with the benchmark. Seminal works in this domain include MultiPath++ \cite{varadarajan2022multipath++}, which combines Top-K trajectories based on confidence scores to integrate multiple model results; ProphNet \cite{wang2023prophnet} utilizes Non-Maximum Suppression (NMS) and QCNet~\cite{zhou2023query} leverages K-means for integrating multiple model results, respectively. Top-K selects the top $K$ trajectories based on confidence scores. NMS combines geometric constraints with the Top-K to balance high scores and diversity by selecting trajectories with high scores while excluding similar ones. K-means clusters trajectories based on the endpoints and performs a weighted average of trajectories within each cluster according to confidence scores. Our proposed temporal ensembling, while similar to model ensembling in its method of expanding the pool of prediction candidates, distinguishes itself through learning-based aggregation. Our proposed aggregation is achieved through mode queries within a DETR-like architecture \cite{carion2020end} and considers traffic context, rather than directly aggregating predictions at the trajectory level.


\section{Proposed Method} 
\label{sec:proposed-method}
This section starts with the baseline utilized and introduces temporal ensembling to address the issue of missing behaviors. It then explains how naively applied trajectory-level aggregation to temporal ensembling is insufficient, leading to the proposal of a learning-based aggregation method. 

\subsection{Baseline Model}
\subsubsection{Input Output Formulation}
QCNet\cite{zhou2023query} was selected as our baseline model due to its state-of-the-art performance. QCNet is the winner of the Argoverse Motion Forecasting 2023 competition. It employs an encoder-decoder architecture with vectorized representation \cite{gao2020vectornet} as input. The agent state comprises the spatial position $p^t_i=(p^t_{i,x}, p^t_{i,y})$, angular position $\theta^t_i$, temporal time $t$, and motion vector $p^t_i - p^{t-1}_i$ for the $i$-th agent at time step $t$, while the HD map contains spatially sampled points and semantic attributes. Input coordinates are set in the world coordinate system while the output trajectories are in the agent-centric coordinate system. This design eliminates redundant input encoding associated with agent-to-agent centric mapping~\cite{zhou2022hivt, wang2023prophnet}, leading to significant savings in computational resources.

\subsubsection{Encoder}
To achieve the world-to-agent coordinate transformation, the encoder utilizes polar representation within each scene element and relative spatial-temporal positional embedding \cite{zhou2023query} between every pair of scene elements. Subsequently, the Fourier transform \cite{mildenhall2021nerf} is applied to scene elements, followed by Multi-Layer Perceptron (MLP) projection to higher-dimensional space. Finally, a factorized attention approach \cite{ngiam2021scene, zhou2022hivt, zhou2023query} is employed to fuse different entities, resulting in the ultimate scene embedding.

\subsubsection{Decoder - DETR-like Architecture}
Recently, trajectory prediction methods\cite{zhou2023query, wang2023prophnet, salzmann2020trajectron++, liu2021multimodal} inspired by DETR\cite{carion2020end} have gained popularity. The DETR-based approach is particularly effective in addressing one-to-many problems, where a single scene embedding corresponds to predicting multiple trajectories. This is achieved through the design of mode queries and the Transformer\cite{vaswani2017attention} structure, involving multiple learnable queries that cross-attend the scene embedding and decode trajectories. For more architectural details, refer to QCNet \cite{zhou2023query}.

\subsubsection{Training Objectives}
Following the approaches \cite{zhou2022hivt, zhou2023query}, we parameterize the future trajectory of the $i$-th agent as a mixture of Laplace distributions:
\begin{equation}
\label{laplace}
    f(\left\{ {p^{t}_{i}} \right\}^{T'}_{t=1} )=\sum^{N}_{n=1}\pi_{i, n} \prod^{T'}_{t=1}Laplace(p^{t}_{i}|\mu^{t}_{i,n}, b^{t}_{i,n}),
\end{equation}
where $\left\{\pi_{i, n} \right\}^{N}_{n=1}$ represents the mixing coefficients, and the Laplace density of the $n$-th mixture component at time step $t$ is defined by the parameters $\mu^{t}_{i,n}$ and $b^{t}_{i,n}$. A classification loss $\mathcal{L}_{cls}$ is utilized to optimize the mixing coefficients. This loss minimizes the negative log-likelihood of Eq. \ref{laplace}. On the other hand, a winner-take-all strategy\cite{lee2016stochastic} is applied to $\mathcal{L}_{traj}$, conducting backpropagation solely on the best-predicted trajectory. The total loss function for end-to-end training is given by:
\begin{equation}
\mathcal{L}_{total} = \mathcal{L}_{traj} + \lambda \mathcal{L}_{cls},
\end{equation}
where $\lambda$ balances regression and classification. Optimization is carried out using the AdamW optimizer\cite{loshchilov2017decoupled} over 64 epochs, with a batch size of 32, a dropout rate of 0.1, and a weight decay coefficient of $1 \times 10^{-4}$. The initial learning rate is set to $5 \times 10^{-4}$ and decayed using the cosine annealing scheduler.

\subsection{Temporal Ensembling}
\subsubsection{Naive Approach with Trajectory-level Aggregation}
In our initial approach to integrating trajectories predicted across multiple time steps, as shown in Fig. \ref{fig:ensemble-comparison}, we utilize the property of large overlaps, depicted in Fig. \ref{fig:streaming-style}. It is found that when evaluating the required horizon at time step $t$, predictions can be made as early as at time step $t-M+1$. This recognition of overlap in prediction horizons allows for the confirmation of predictions across frames, compensating for errors in single-frame predictions.  Initially, trajectories for each nearby time step are predicted. Subsequently, overlapping segments with the target horizon at time step $t$ are sliced, ensuring within the same time range. At this point, we achieve $M \times N$ trajectories, as shown in the middle panel of Fig. \ref{fig:ensemble-comparison}, with $N$ being the number of trajectories predicted per time step and $M$ the number of frames to integrate. A final aggregation stage then merges these into the ultimate $M$ trajectories, illustrated on the rightmost side of Fig. \ref{fig:ensemble-comparison}. For thorough evaluation, we set $M=10$ and $N=6$. 
\begin{table}[!ht]
    \caption{Performance comparison of temporal ensembling with trajectory-level aggregation techniques on the Argoverse 2 validation set. The baseline observes 50 frames, predicts 60 future frames, and evaluates at intervals [10,60). For a fair comparison, single-frame predictions at $t=10$ are made to match temporal ensembling by including the most recent agent states.}
    \label{tab:issue-of-different-aggregation}
    \scriptsize	
    \begin{tabularx}{\columnwidth}{l >{\hsize=.3\hsize}Y >{\hsize=.3\hsize}Y > {\hsize=.55\hsize}Y}
        \toprule
        Method &  minADE & minFDE & MissRate \\
        \midrule
        {QCNet \cite{zhou2023query}}  & \textbf{0.50} & \textbf{0.99} & 10.73\% \\
        {QCNet \cite{zhou2023query} + TempEns w/ Top-K}  & 0.72 & 1.65 & 24.80\% \\
        {QCNet \cite{zhou2023query} + TempEns w/ NMS}  & 0.64 & 1.19 & 12.27\% \\
        {QCNet \cite{zhou2023query} + TempEns w/ K-means}  & 0.51 & 1.01 & \textbf{10.72\%} \\
        \bottomrule
    \end{tabularx}
    \vspace*{-0.6em}
\end{table}

\begin{figure}[t]
\centering
\includegraphics[height=4.3cm]{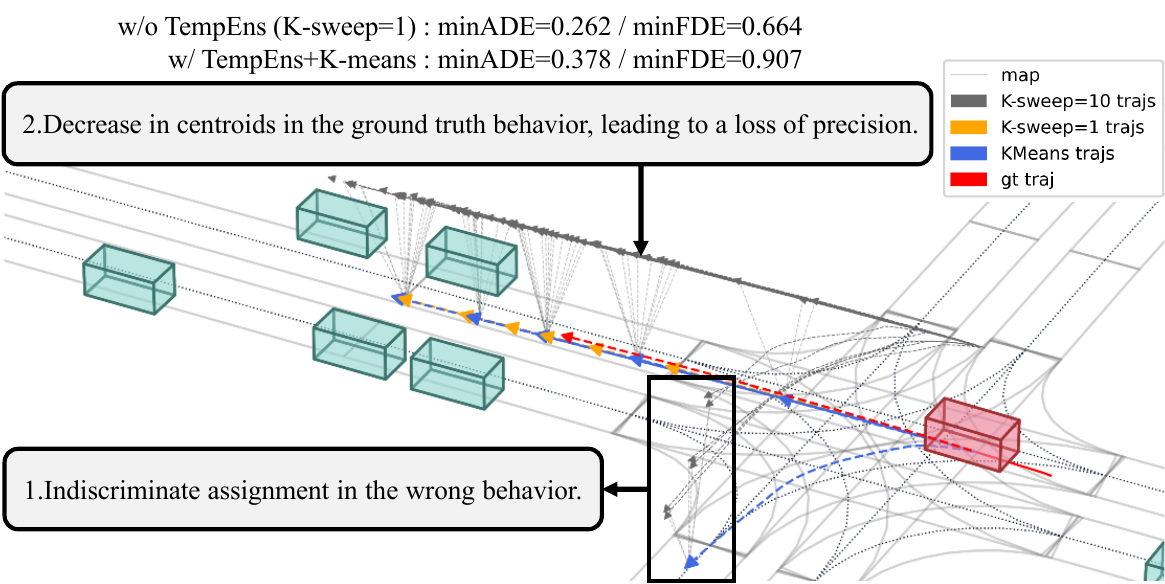}
\caption{This highlights the precision-diversity trade-off in trajectory-level aggregation using K-means. Gray trajectories represent all trajectories within the sliding window. Orange trajectories depict single-frame approach predictions, while blue trajectories demonstrate the integration of multiple-frame predictions aggregated from the gray ones.}
\label{subf:issue-analysis}
\vspace*{-0.6em}

\end{figure}
In our initial experiment, which involved integrating trajectory-level aggregation into temporal ensembling, as illustrated in Table \ref{tab:issue-of-different-aggregation}, we discovered that trajectory-level aggregation was less effective than anticipated. Specifically, predictions based on a single frame outperformed those derived from multiple frames. Our analysis revealed a pattern across multiple frames predictions where high-scoring trajectories often exhibit similarity. This phenomenon adversely affects when applying selection manners (Top-K and NMS), leading to a decrease in diversity and consequently degrading performance. Further analysis presented in Fig. \ref{subf:issue-analysis} reveals two key insights regarding the clustering manner (K-means). First, integrating trajectories through distance-based clustering enhanced diversity by preserving geometrical variations, including both straight and left-turn trajectories. Second, however, this approach sometimes reduced accuracy, particularly when the ground truth was a straight trajectory. K-means' indiscriminate assignment of candidate trajectories to different behaviors, without re-considering traffic context, often resulted in inaccurate aggregation.

\subsubsection{Learning-based Aggregation}
\begin{figure*}[ht]
\includegraphics[width=\textwidth]{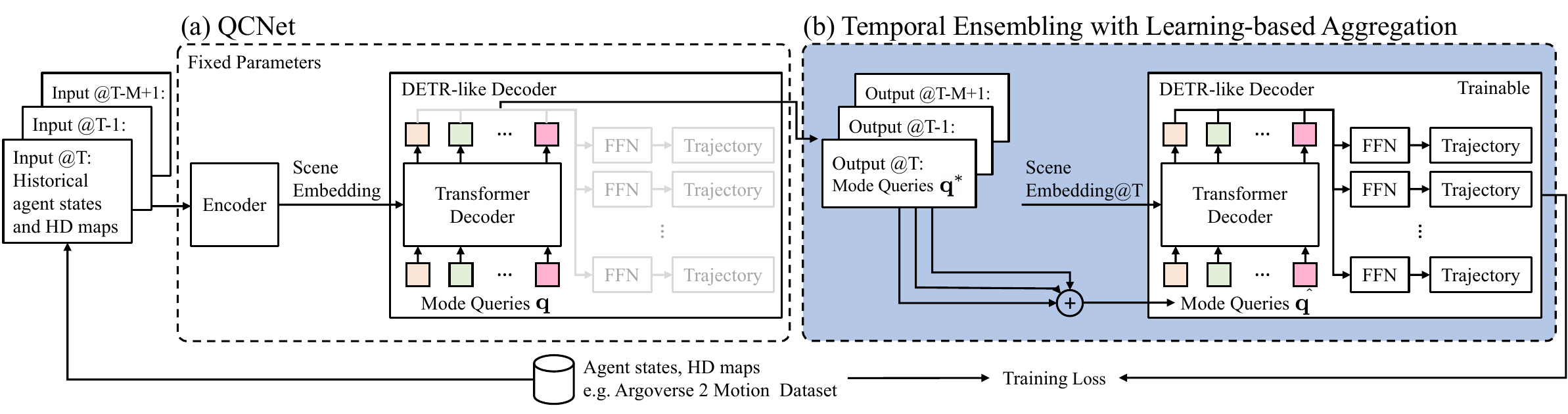}
\caption{
\textbf{Overall Pipeline of Temporal Ensembling with Learning-based Aggregation - } The architecture consists of two main blocks. \textbf{Block (a)} represents the baseline model, QCNet, from which we leverage the predicted mode queries (not the final trajectories). \textbf{Block (b)} depicts our proposed method. It takes predictions of mode queries from nearby frames as input. Element-wise addition is used to aggregate historical mode queries. A transformer decoder then fuses the aggregated mode queries with scene embedding at time step T. Finally, each feed-forward network (FFN) predicts the final trajectory.}
\label{fig:training-cycle}
\vspace*{-0.6em}

\end{figure*}
To address the issues associated with trajectory-level aggregation, we propose integrating the aggregation process into the model pipeline, with an emphasis on considering the traffic context. We propose leveraging mode query designs in a DETR-like decoder\cite{zhou2023query}, marking a shift toward using vectors formed by the transformer decoder with scene embeddings, as opposed to directly using trajectories. This vector encapsulates potential driving intentions, which are then transformed into trajectories. By focusing on this high-dimensional space, we identify an ideal opportunity for learning-based aggregation. This technique allows us to combine driving intentions across time, effectively addressing behaviors that are missed in predictions based on a single frame. As depicted in Fig. \ref{fig:training-cycle}, the mode queries we utilized are expressed as:
\begin{equation}
\mathbf{q}^{*}_{nt} = \texttt{TransformerDecoder}(\mathbf{q}_{nt}, \mathbf{s}_{t}),
\end{equation}
where $\mathbf{q}_{nt}$ denotes the $n$-th initial mode query at time step $t$, with $n \in {1, 2, \dots, N}$ and $N$ representing the total number of mode queries. $\mathbf{s}_{t}$ represents the scene embedding generated by the encoder at time step $t$. Each $\mathbf{q}^{*}_{nt}$, the output of the transformer decoder, is tasked with predicting final trajectories and representing driving intentions within the spatial-temporal scene context. We collect historical mode queries from nearby frame predictions, each generating $N$ mode queries, resulting in $M \times N$ mode queries, where $M$ is the number of nearby frames. Through element-wise addition, denoted as
\begin{equation}
\mathbf{q}^{\hat{}}_{nt} = \mathbf{q}^{*}_{nt} + \mathbf{q}^{*}_{n{t-1}} + \dots + \mathbf{q}^{*}_{n{t-M+1}},
\end{equation}
we combine the corresponding positions of the N mode queries $\mathbf{q}^{*}_{nt}$ to obtain the aggregated mode query $\mathbf{q}^{\hat{}}_{nt}$. The slight time offset between the N mode queries $\mathbf{q}^{*}_{nt}$ for nearby frames, there is a need to re-utilize the scene embedding $\mathbf{s}_{t}$ to refine $\mathbf{q}^{\hat{}}_{nt}$. We employ a baseline transformer decoder\cite{zhou2023query} to fuse $\mathbf{q}^{\hat{}}_{nt}$ with the scene embedding $\mathbf{s}_{t}$. This creates a context-aware mode query $\mathbf{\tilde{q}}_{nt}$. Feed Forward Networks (FFNs) then convert the final state of $N$ mode queries $\mathbf{\tilde{q}}_{nt}$ into $N$ trajectories as follows:
\begin{equation}
\mathbf{\tilde{q}}_{nt} = \texttt{TransformerDecoder}(\mathbf{q}^{\hat{}}_{nt}, \mathbf{s}_{t}),
\end{equation}
\begin{equation}
\mathbf{trajectory}_{nt} = \texttt{FFN}(\mathbf{\tilde{q}}_{nt}).
\end{equation}

\subsubsection{Two Operations within Learning-based Aggregation}
\begin{table}[!ht]
    \caption{This table presents the results of two aggregation operations within our learning-based aggregation method on the Argoverse 2 validation set. The baseline observes 50 past frames to predict 60 future steps (evaluated at intervals [10, 60)).}
    \label{tab:comparison-with-different-fusion}
    \scriptsize	
    \begin{tabularx}{\columnwidth}{l >{\hsize=.3\hsize}Y >{\hsize=.3\hsize}Y > {\hsize=.55\hsize}Y}
        \toprule
        Method &  minADE & minFDE & MissRate \\
        \midrule
        {LearnAgg w/ CrossAttn}  & 0.48 & 0.95 & 9.72\% \\
        {LearnAgg w/ ElementWiseAdd}  & \textbf{0.48} & \textbf{0.94} & \textbf{9.57\%} \\
        \bottomrule
    \end{tabularx}
    \vspace*{-1.2em}
\end{table}

In learning-based aggregation, the most straightforward operation involves using a cross-attention mechanism \cite{vaswani2017attention}, where queries are the mode queries from the current frame, and keys and values are the anticipated integrated mode queries from nearby frames. Cross-attention aims to harness the capability of attention mechanisms to identify crucial behavioral features that might be missed in the current frame. Its effectiveness is shown in Table \ref{tab:comparison-with-different-fusion}, with improvements in all three main metrics. However, by leveraging the DETR-like structure \cite{carion2020end} and the design of mode queries, we further explore a simpler operation: element-wise addition of mode queries. This approach is motivated by the desire to capture missing behaviors from nearby frame predictions. At a high level, it aims to encompass driving behaviors by incorporating the right behaviors into the current prediction. By overlapping the high-dimensional vectors, the full spectrum of driving intentions from nearby frames can be covered, which yields even better results. This improvement is attributed to the additive nature of mode queries.

\subsubsection{Training Pipeline}
Our proposed training pipeline for temporal ensembling with learning-based aggregation, as depicted in Fig. \ref{fig:training-cycle}, comprises two blocks. In  Fig. \ref{fig:training-cycle}-(a), a fully trained base model is prepared, and its parameters are then frozen. Subsequently, predictions are made on continuous data. The mode queries across multiple frames produced by the base model are collected and used as the proposed inputs. These are then fed into the newly added decoder \cite{zhou2023query}, as shown in Fig. \ref{fig:training-cycle}-(b), resulting in the final set of N trajectories. The training loss for our proposed method is aligned with the loss of 
the base model. The parameters of the newly added decoder are fine-tuned, starting with an initial learning rate of 2.5e-4, which is half of the base model. The AdamW optimizer\cite{loshchilov2017decoupled}, along with a cosine annealing scheduler, is employed for a total duration of 8 training epochs.
\section{Experiments}
\label{sec:experiments}

In this section, 
we commence by presenting comprehensive information about the dataset and metrics utilized. Subsequent comparisons with the state-of-the-art methods and qualitative results are provided. Additionally, the combined effect of model ensembling and the proposed temporal ensembling is explored to demonstrate their extensibility and effectiveness.

\subsection{Experimental Setup}
\begin{figure}[t]
\includegraphics[height=2.75cm]{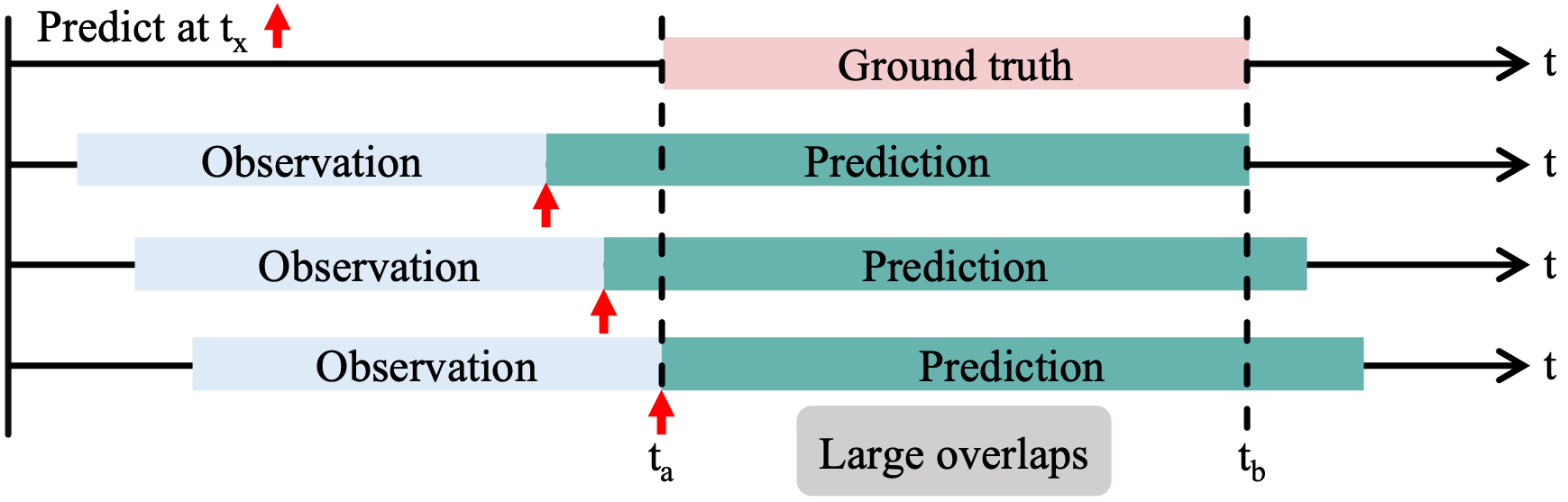}
\caption{Streaming-style formulation - Predictions exhibit a high degree of overlap in continuous datasets in the streaming-style paradigm. The overlapping time ranges, shown by the two dashed lines, offer an opportunity to exploit this property.}
\label{fig:streaming-style}
\vspace*{-0.6em}
\end{figure}
\label{experimental_setup}
\subsubsection{Dataset and Streaming-Style Formulation}
Our method is evaluated 
using the Argoverse 2 Dataset~\cite{wilson2021argoverse} which is widely recognized for its diverse and challenging scenarios. This dataset contains 250K non-overlapping scenarios. Trajectories are sampled at 10Hz, with an observation window of (-50, 0] frames and a prediction horizon of (0, 60] frames. Each scenario is associated with a local map region, including geometric properties such as lane centerlines, boundaries, crosswalks, and road markers. We redefine the selected dataset \cite{wilson2021argoverse} to exhibit streaming-style characteristics by transforming each data segment. Rather than predicting each snapshot once within a fixed-length segment, our approach employs a sliding-window temporal arrangement, as depicted in Fig. \ref{fig:streaming-style}, to introduce streaming characteristics. This arrangement can be achieved through two approaches within the selected dataset \cite{wilson2021argoverse}: One approach maintains the original configuration of the model while shortening the evaluation length. The other conforms to the benchmark evaluation setting by adjusting the model to observe shorter periods and predict beyond the evaluation length. Experiments are conducted in both settings to demonstrate the effectiveness of the proposed method.

\subsubsection{Metrics}
To evaluate prediction performance, standard metrics are employed: minimum final displacement error (minFDE), minimum average displacement error (minADE), and miss rate, aiming for lower values across up to N=6 predicted trajectories per agent. \textbf{minFDE} measures the $\ell^{2}$ distance between the endpoints of ground truth trajectory and predicted trajectory $\hat{N}$, where $\hat{N}$ is the trajectory index with the minimum endpoint error among $N$ predicted trajectories. \textbf{minADE} calculates the average $\ell^{2}$ distance between predicted trajectory $\hat{N}$ and ground truth over all time steps. \textbf{Miss Rate} reflects the proportion of scenarios where the distance between the ground truth endpoint and the best-predicted endpoint exceeds 2.0 meters.


\begin{figure*}[ht]
\centering
\includegraphics[width=\textwidth]{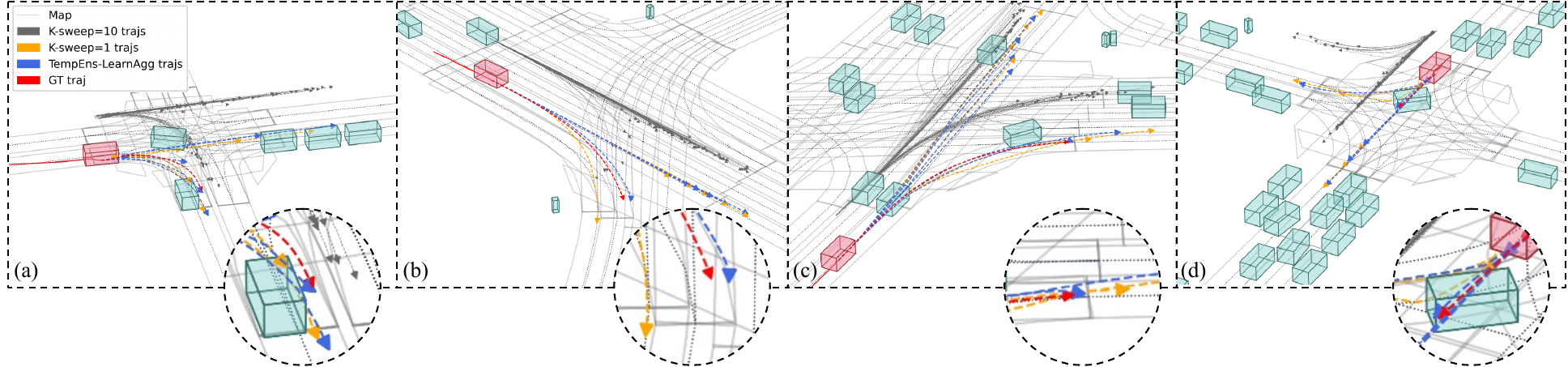}

\caption{Qualitative Results - Demonstrate the results of employing temporal ensembling with learning-based aggregation. The trajectories are color-coded: red for ground truth, orange for single-frame predictions, and blue for predictions using temporal ensembling with learning-based aggregation. Gray trajectories depict recent frame predictions, aiding in understanding the aggregation process.}
\label{subf:cases}
\vspace*{-0.6em}

\end{figure*}

\subsection{Quantitative Results}
\begin{table}[!ht]
    \caption{Quantitative Results I - This table presents the performance of QCNet using temporal ensembling with learning-based aggregation on the Argoverse 2 validation set. The baseline observes 50 past frames to predict 60 future steps (evaluated at intervals [10, 60)). For a fair comparison, single-frame predictions at $t=10$ are made to match temporal ensembling by including the most recent agent states.}
    \label{tab:comparison-with-different-aggregation}
    \scriptsize	
    \begin{tabularx}{\columnwidth}{l >{\centering\arraybackslash\hsize=.3\hsize}X >{\centering\arraybackslash\hsize=.3\hsize}X >{\centering\arraybackslash\hsize=.5\hsize}X}
    \toprule
    Method & minADE & minFDE & MissRate \\
    \midrule
    {QCNet \cite{zhou2023query}}  & 0.50 & 0.99 & 10.73\% \\
    \midrule
    \multirow{2}{*}{TempEns w/ LearnAgg + QCNet \cite{zhou2023query}}  & \textbf{0.48} & \textbf{0.94} & \textbf{9.57\%} \\
    & \textbf{(4.0\%↓)} & \textbf{(5.0\%↓)} & \textbf{(1.16\%↓)} \\
    \bottomrule
\end{tabularx}
\end{table}
We compare our proposed pipeline, Temporal Ensembling, with the strongest baseline, QCNet\cite{zhou2023query}, on the Argoverse 2 dataset\cite{wilson2021argoverse}. The results in Table \ref{tab:issue-of-different-aggregation} reveal that trajectory-level aggregations using KMeans, NMS, and TopK exhibit performance inferior to QCNet\cite{zhou2023query}, contrary to our initial motivation. However, the temporal ensembling with learning-based aggregation 
shown in Table \ref{tab:comparison-with-different-aggregation} surpasses all trajectory-level aggregation methods and even outperforms QCNet\cite{zhou2023query} in three metrics. Specifically, it achieves a 4\% improvement in minADE, a 5\% improvement in minFDE, and a 1.16\% improvement in Miss Rate. These results align with our motivation, highlighting the feasibility of integrating predictions across multiple frames.

\begin{table}[!ht]
    \caption{Quantitative Results of Different Base Model Configurations - We investigated the effect of base model settings on performance. Both models were trained on a GeForce RTX 4090, differing only in the observation window and prediction horizon. The (40/70) base model, observing 40 frames and predicting 70 (evaluated at intervals [0, 60), enabled further temporal ensembling.}
    \label{tab:comparison-with-5060}
    \scriptsize	
    \begin{tabularx}{\columnwidth}{X >{\hsize=.3\hsize}Y >{\hsize=.3\hsize}Y > {\hsize=.55\hsize}Y}
        \toprule
        Method &  minADE & minFDE & MissRate \\
        \midrule
        {QCNet\cite{zhou2023query} - Naive (50/60)}  & \textbf{0.65} & \textbf{1.27} & \textbf{0.16} \\
        {QCNet \cite{zhou2023query} - (40/70)}  & 0.68 & 1.32 & 0.17 \\
        \bottomrule
    \end{tabularx}
    \vspace*{-0.6em}

\end{table}
\begin{table}[!ht]
    \caption{Quantitative Results II - The table compares our method's performance on the Argoverse 2 test set to SOTA approaches. For the official benchmark, evaluation is conducted at the [0, 60) interval. * denotes the (40/70) base model. \# indicates methods without publicly available code. All results exclude model ensembling for a fair comparison.}
    \label{tab:comparison-with-sota}
    \scriptsize	
    \begin{tabularx}{\columnwidth}{l >{\hsize=.3\hsize}Y >{\hsize=.3\hsize}Y > {\hsize=.5\hsize}Y}
        \toprule
        Method &  minADE & minFDE & MissRate \\
        \midrule
        {GANet\# (ICRA 2023)\cite{wang2023ganet}}  & 0.69 & 1.33 & 0.18 \\
        {ProphNet\# (CVPR 2023) \cite{wang2023prophnet}}  & 0.65 & 1.31 & 0.17 \\
        {QCNet* (CVPR 2023)\cite{zhou2023query} }  & 0.68 & 1.32 & 0.17 \\
        \midrule
        {TempEns w/ LearnAgg + QCNet* \cite{zhou2023query}}  & \textbf{0.65} & \textbf{1.27}& \textbf{0.16}\\
        \bottomrule
    \end{tabularx}
\end{table}
In the second quantitative result as shown in Table \ref{tab:comparison-with-sota}, we evaluate our performance against the state-of-the-art methods on the Argoverse 2 test set~\cite{wilson2021argoverse}. There are two key findings. Firstly, reducing the observation window size can lead to a degradation in model performance (Table \ref{tab:comparison-with-5060}). Secondly, in subsequent experiments, incorporating our proposed approach to the base model with shorter observation periods demonstrates observable improvements across all evaluation metrics, reaffirming the effectiveness of our method.

\subsection{Analysis of Computational Overhead}
\begin{table}[ht]
\centering
\caption{Computational Overhead - We measured inference time using a GeForce RTX 4090 in the \textbf{densest} traffic scene involving \textbf{190} agents and \textbf{169} map polygons.}
\label{computational-overhead}
\begin{tabular}{m{2.cm}m{4cm}<{\centering} }

 \toprule
 Method & Online Inference Time\\ 
 \midrule
 w/o TempEns & 80 $\pm$ 1 (ms) \\ 
 w/ TempEns & 108 $\pm$ 1 (ms) \\
 \bottomrule
\end{tabular}
\end{table}
During training, our approach adopts a pre-trained QCNet \cite{zhou2023query} as the base model and only fine-tunes the newly added decoder. Practically, the resources needed for fine-tuning are significantly fewer compared to retraining a base model. 
During inference, temporal ensembling can be achieved with minimal computational resources, as it only requires a single forward pass. Historical mode queries are temporarily cached as the model progresses, enabling temporal ensembling by directly accessing recent data at the current time step. We also conducted experiments on inference time 
as shown in Table \ref{computational-overhead} to demonstrate that a slight increase in computational resources leads to significant performance improvements.

\subsection{Qualitative Results}
In Fig. \ref{subf:cases}, we showcase qualitative results from the Argoverse 2 validation set~\cite{wilson2021argoverse}. These cases highlight instances where the baseline method \cite{zhou2023query} (orange trajectories) failed, particularly in cases where the endpoint accuracy of the best trajectory exceeds a 2-meter error. Conversely, our approach (blue trajectories) successfully predicts the exact behaviors.

\subsection{Dual Ensembling with Varied Aggregation Techniques}
\label{ablation_study}
\begin{table}[!ht]
    \caption{Exploring Model and Temporal Ensembling with Different Aggregation on the Argoverse 2 validation set. The baseline observes 50 past frames to predict 60 future steps, evaluated at intervals [10, 60).}
    \label{tab:both-ensembling-v2}
    \scriptsize
    \begin{tabular}{m{4cm}m{1cm}<{\centering}m{1cm}<{\centering}m{1cm}<{\centering}}
    \toprule
    Method & minADE & minFDE & MissRate \\ 
    \midrule
    Baseline & 0.50 & 0.99 & 10.73\% \\ 
    + ModelEns w/ TrajAgg & 0.50 & 0.98 & 10.20\% \\ 
    + ModelEns w/ LearnAgg & 0.49 & 0.96 & 9.84\% \\ 
    + TempEns w/ LearnAgg & 0.48 & 0.94 & 9.57\% \\ 
    + ModelEns \& TempEns w/ LearnAgg & \textbf{0.48} & \textbf{0.94} & \textbf{9.52\%} \\ 
    \bottomrule
\end{tabular}

\end{table}
We explore the combined effects of model and temporal ensembling and their applicability when integrated. The number of models in model ensembling is determined by computational resources. For our experiment, we train three model instances with varying epochs for model ensembling. In the initial model ensembling with trajectory-level aggregation experiment (Table \ref{tab:both-ensembling-v2}'s second row), predictions from three separate models are merged and combined using K-means, resulting in improvements in miss rate and minFDE, as expected. In the third row, we replaced the trajectory-level aggregation with our proposed learning-based aggregation. We observed that it outperformed trajectory-level aggregation, which we attribute to its effectiveness in integrating features from different models and then re-considering surrounding traffic information, thereby improving prediction accuracy. The fourth row of Table \ref{tab:both-ensembling-v2} represents the results of the pipeline proposed in this paper. We conclude that both components are crucial. \textbf{Temporal Ensembling} contributes diversity, while \textbf{Learning-based Aggregation} enhances precision. In the final row of Table \ref{tab:both-ensembling-v2}, we employ a dual approach by first averaging the mode queries $\mathbf{q}^{*}_{nt}$ across models during model ensembling and then applying the proposed temporal ensembling pipeline from nearby frames, which performs best across all three major metrics. Combining both pipelines yielded further performance enhancements.

\subsection{Alternative Base Model with DETR-like Architecture}
\label{ablation_study_v2}
\begin{table}[!ht]
    \caption{Ablation Study on Alternative Base Models - This table showcases the performance of mmTrans using temporal ensembling with learning-based aggregation on the Argoverse 1 validation set. The base model uses 20 past frames to forecast 30 future steps, evaluated at intervals of [4, 30). To ensure a fair comparison, single-frame prediction at $t=4$ includes the most recent agent states to align with our approach.}
    \label{tab:comparison-with-mmTrans-v2}
    \scriptsize	
    \begin{tabular}{m{3.5cm}m{1cm}<{\centering}m{1cm}<{\centering}m{1.5cm}<{\centering}}
    \toprule
    Method & minADE & minFDE & MissRate \\ 
    \midrule
    mmTrans \cite{liu2021multimodal} & 0.62 & 0.90 & 7.38\% \\ 
    TempEns w/ LearnAgg + mmTrans & \textbf{0.61} & \textbf{0.88} & \textbf{6.82\%} \\
    \bottomrule
\end{tabular}

\end{table}
We applied our proposed meta-algorithm, Temporal Ensembling with Learning-based Aggregation, to another well-renowned DETR-like model, mmTrans\cite{carion2020end}, to assess its impact on performance. Our results showed significant improvements in all three major performance indicators, further validating the effectiveness of our approach.
\section{Conclusion and Future Works}
\label{chatper:conclusion}
In this paper, we introduce a simple yet effective meta-algorithm, Temporal Ensembling with Learning-based Aggregation, to address the challenge of missing behaviors, effectively compensating single-frame predictions from multiple frames. Experimental results validate the effectiveness of the proposed approach, aiming to spark interest in exploring motion forecasting within a realistic streaming setting. A limitation of our method is its dependence on the assumption that nearby frames contain accurate predictions, making it less effective when predictions consecutively fail. For future work, we aim to comprehensively address this 
issue affecting safety by developing solutions that ensure robustness and accuracy in motion prediction across various scenarios and over time.

\bibliographystyle{IEEEtran}

\end{document}